\documentclass[runningheads]{llncs}

\usepackage{eccv}

\usepackage{eccvabbrv}

\usepackage{graphicx}
\usepackage{booktabs}
\usepackage{adjustbox}

\usepackage[accsupp]{axessibility}  %

\usepackage{hyperref}

\usepackage{orcidlink}
\usepackage{multirow}

\usepackage{mathtools}

\def\eg{\emph{e.g}.} 
\def\ie{\emph{i.e}.}

\def\etal{\emph{et al}.~}
\newcommand{\quotes}[1]{``#1''}

\begin{document}

\title{Rethinking Directional Parameterization in \\ Neural Implicit Surface Reconstruction}

\titlerunning{Rethinking directional parameterization}

\author{Zijie Jiang\inst{1}\thanks{Equal contribution. Work done while Zijie Jiang interned at Preferred Networks, Inc.}\orcidlink{0000-0002-3438-7480} \and
Tianhan Xu\inst{2}$^\star$\orcidlink{0009-0006-5085-6773} \and
Hiroharu Kato\inst{2}\orcidlink{0000-0003-0920-3181}}

\authorrunning{Jiang and Xu et al.}

\institute{Tokyo Institute of Technology \and
Preferred Networks, Inc.
\\
\email{zjiang@ok.sc.e.titech.ac.jp, \{xutianhan,hkato\}@preferred.jp}}

\maketitle
\begin{abstract}
  Multi-view 3D surface reconstruction using neural implicit representations has made notable progress by modeling the geometry and view-dependent radiance fields within a unified framework.
  However, their effectiveness in reconstructing objects with specular or complex surfaces is typically biased by the directional parameterization used in their view-dependent radiance network.
  {\it Viewing direction} and {\it reflection direction} are the two most commonly used directional parameterizations but have their own limitations.
  Typically, utilizing the viewing direction usually struggles to correctly decouple the geometry and appearance of objects with highly specular surfaces, while using the reflection direction tends to yield overly smooth reconstructions for concave or complex structures.
  In this paper, we analyze their failed cases in detail and propose a novel hybrid directional parameterization to address their limitations in a unified form.
  Extensive experiments demonstrate the proposed hybrid directional parameterization consistently delivered satisfactory results in reconstructing objects with a wide variety of materials, geometry and appearance, whereas using other directional parameterizations faces challenges in reconstructing certain objects.
  Moreover, the proposed hybrid directional parameterization is nearly parameter-free and can be effortlessly applied in any existing neural surface reconstruction method.
\end{abstract}

\section{Introduction}
\label{sec:intro}

Recent advances in neural fields~\cite{xie2022neural} have revolutionized the field of 3D geometry reconstruction from multi-view images~\cite{wang2021neus,unisurf,idr}. Compared to traditional multi-view stereo methods~\cite{furukawa2015multi}, a key advantage of using neural fields is their ability to model and optimize complex light reflections on object surfaces, which improves robustness to specular or non-Lambertian surfaces. Since a radiance network is trained jointly with a geometry network via differentiable rendering, the radiance network is able to take full advantage of geometry and view-dependent information such as viewing direction to decouple the geometry and appearance.

\begin{figure}[tb]
  \centering
  \includegraphics[width=1\linewidth]{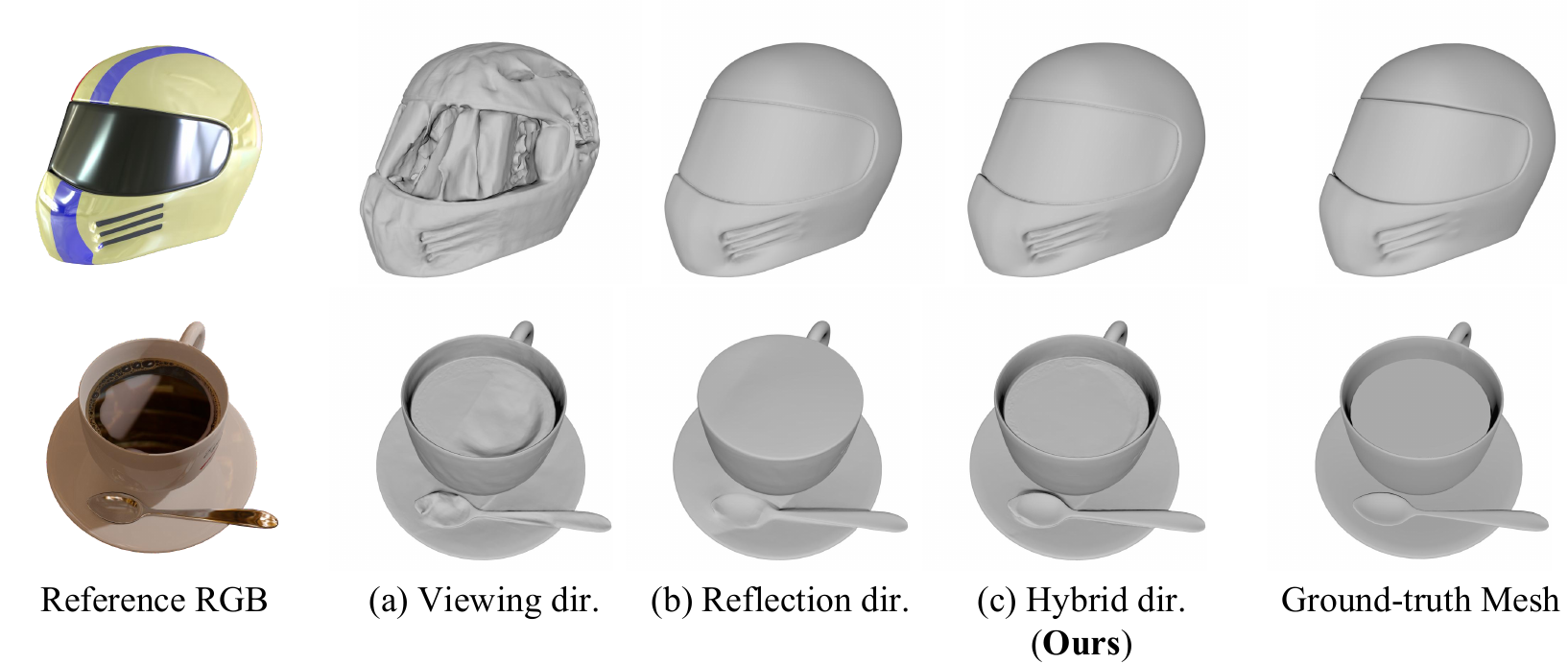}
  \caption{\textbf{Comparison of reconstruction results using different directional parameterizations in ~\cite{cai2023neuda}.}
  Using the viewing directional parameterization fails to reconstruct objects with specular surfaces, while using the reflection directional parameterization tends to result in incorrect geometry for objects with concave structures.
  Using our proposed hybrid parametrization shows consistently satisfactory reconstruction results compared to the existing parameterizations.
  }
  \label{fig:intro_1}
\end{figure}

Since the decomposition of geometry and appearance is inherently an ill-posed problem~\cite{zhang2020nerf}\footnote{If the capacity of the radiance network is sufficiently high, even for a fixed geometry such as a sphere, the image reconstruction loss can converge to zero~\cite{zhang2020nerf}.}, it is essential to introduce an appropriate inductive bias into the radiance network to facilitate correct decomposition. For example, simply inputting the viewing direction to the radiance network sometimes leads to incorrect reconstructions of highly specular objects (Fig.~\ref{fig:intro_1} (a)). To address this issue, the recent study~\cite{verbin2022ref} suggests that providing the reflection direction of rays considering the learned local surface geometry (\ie, normals) would result in a simpler radiance field to learn and facilitate correct decomposition, especially for specular objects.
Some subsequent studies have shown that adopting this parameterization results in satisfactory reconstructions of highly specular objects~\cite{ge2023ref, liu2023nero, liang2023envidr} and also non-reflecting objects~\cite{ge2023ref}. However, although this parameterization seems promising (Fig.~\ref{fig:intro_1} (b) top), we observed that it still does not work well when reconstructing objects with complex structures such as concave surfaces (Fig.~\ref{fig:intro_1} (b) bottom).
Interestingly, while using viewing direction fails to recover specular surfaces, it succeeds in recovering such concave structures (Fig.~\ref{fig:intro_1} (a) bottom).
Based on these, we pose the question:
\textbf{Is there a unified directional parameterization that can effectively handle both specular surfaces and complex structures?}

In this paper, to answer the question, we start by studying the limitations of existing directional parameterizations and provide the first in-depth analysis of why using reflection direction may fail to reconstruct objects with concave or complex structures.
As mentioned above, one important feature of the reflection directional parameterization is its dependency on the learnable geometry (\ie, normals).
Our insight into this fact is that for sampled 3D points along a sampled ray during training, while utilizing reflection directions of sampled points near the intersecting surface is helpful in decoupling geometry and appearance, the use of reflection directions of sampled points elsewhere may hinder the accurate reconstruction of geometry.
Specifically, we identified that the cause of this issue lies in the sampled points beyond a certain distance from the intersecting surface during the optimization. As we will discuss in Sec.~\ref{subsec:3_2}, the normals of these points computed from learnable geometry may either be associated with surfaces unrelated to the intersection or introduce high-frequency variations in the input of the radiance network, which hinders the proper updating of geometry and radiance.
This hindrance is particularly noticeable in the reconstruction of concave surfaces.
The analysis also indirectly explains the advantage of using viewing directional parameterization in reconstructing concave or complex structures, as it is non-learnable and remains constant for all sampled points along a sampled ray.

Based on our insights and analysis, we propose a novel hybrid directional parameterization to address the limitations of existing directional parameterizations in a unified form.
The main idea is to utilize the predicted distance to the surface for each sampled point: if the distance is small, indicating proximity to the intersecting surface, it is preferable to use its reflection direction as the directional parameterization, which facilitates the correct decomposition of geometry and appearance; in other cases, we instead utilize the non-learnable viewing direction to avoid the optimization issue mentioned above.
The proposed hybrid directional parameterization smoothly transitions from the reflection direction to the viewing direction based on the predicted distance to the surface.
The rate of this directional transition is controlled by a learnable scaling factor.
We integrate the proposed hybrid directional parameterization into different methods and conduct extensive experiments to validate its effectiveness across various datasets.
Compared to the existing viewing and reflection directional parameterizations, our proposed hybrid directional parameterization achieves consistently satisfactory results, as shown in Fig.~\ref{fig:intro_1} (c).
Moreover, the proposed hybrid directional parameterization is nearly parameter-free and can be effortlessly applied in any existing neural implicit surface reconstruction method.

To summarize, our contributions are as follows:

\begin{itemize}
  \item We are the first to provide an in-depth analysis of why the use of reflection directional parameterization tends to reconstruct incorrect geometry of objects with concave or complex structures.
  \item We propose a novel hybrid directional parameterization
  to address the limitations of existing directional parameterizations, which can effectively handle both specular surfaces and complex structures in a unified form.
  \item We validate the effectiveness of the proposed directional parameterization across different methods on various datasets, and demonstrate an overall superiority compared to existing directional parameterizations.
\end{itemize}

\section{Related Work}
\label{sec:related_work}

\subsection{Neural Implicit Surface Reconstruction}

With the significant advancement of neural implicit representations~\cite{park2019deepsdf,peng2020convolutional,chen2019learning,mescheder2019occupancy,muller2022instant,mildenhall2020nerf,niemeyer2019occupancy,yu2021pixelnerf} and differentiable rendering~\cite{kato2020differentiable, niemeyer2020differentiable,yan2016perspective,tulsiani2017multi,li2018differentiable,ravi2020accelerating,mildenhall2020nerf} in recent years, neural implicit surface reconstruction methods~\cite{DVR, idr, wang2021neus, yariv2021volume} have emerged as a promising alternative to traditional methods in the field of image-based 3D surface reconstruction.
Compared to traditional multi-view stereo methods, neural implicit surface methods employ a more flexible and representative implicit shape representation, such as the signed distance function (SDF), and introduce the directional radiance field~\cite{mildenhall2020nerf} to model the view-dependent appearance of the target object.
Subsequent works on neural implicit surface reconstruction have mostly built upon the foundation of NeuS~\cite{wang2021neus} or VolSDF~\cite{yariv2021volume}, and made improvements for specific scenarios including fast rendering~\cite{wu2022voxurf,wang2023neus2}, sparse-view inputs~\cite{long2022sparseneus}, large-scale scenes~\cite{li2023neuralangelo, Yu2022MonoSDF} and more.

\subsection{Reconstruction for Specular Objects}
Despite neural implicit surface reconstruction methods consider the view-dependent appearance of objects by incorporating the directional radiance field, reconstruction for specular objects remains a challenging task due to its significant shape-radiance ambiguity.
Several works~\cite{liu2023nero, neural-incident-light-field, zhang2023neilf++, srinivasan2021nerv, zhang2021nerfactor} model the view-dependent appearance of specular objects by incorporating additional physical components (\eg, materials and lighting), with the hope of jointly recovering both the geometric shape and physical parameters.
However, as the number of unknown parameters increases, these works typically rely on a good initial geometry to ensure favorable results, leading to a chicken-and-egg problem.
On the other hand, Ref-NeRF~\cite{verbin2022ref} was the first to propose the use of a radiance field conditioned on reflection direction to better decouple the geometry and radiance of specular objects.
Subsequent neural implicit surface reconstruction methods targeting specular objects~\cite{ge2023ref, liang2023envidr, liu2023nero} have also adopted this directional parameterization.
However, these methods often fail to correctly recover the geometry of objects with concave or complex structures.

\section{Method}
\label{sec:method}

In this section, we introduce a novel directional parameterization for neural implicit surface reconstruction methods.
We first give a brief review on neural implicit surface reconstruction and its commonly used directional parameterizations, including viewing and reflection directions (Sec.~\ref{subsec:3_1}).
In the next, we summarize the observed failed reconstructions when utilizing existing directional parameterizations and conduct a detailed analysis of their causes (Sec.~\ref{subsec:3_2}).
Finally, a hybrid directional parameterization is proposed based on our analysis, which can effectively address the limitations of existing directional parameterizations in a unified form (Sec.~\ref{subsec:3_3}).

\subsection{Preliminaries of Neural Implicit Surface Reconstruction}
\label{subsec:3_1}

Neural implicit surface reconstruction methods~\cite{idr,wang2021neus, cai2023neuda, Fu2022GeoNeus} typically represent the 3D object as the combination of a signed distance function (SDF) network $\Theta_f$, which maps a spatial position $\mathbf{x} \in \mathbb{R}^3$ to its SDF value $f(\mathbf{x}) \in \mathbb{R}$, and a directional radiance network $\Theta_c$, which maps a spatial position $\mathbf{x} \in \mathbb{R}^3$ and a directional vector $\mathbf{d} \in \mathbb{S}^2$ to the view-dependent color at $\mathbf{x}$.
The surface $\mathcal{S}$ of the object is extracted as the zero-level set of $f(\mathbf{x})$.
To enable the joint optimization of the SDF network $\Theta_f$ and radiance network $\Theta_c$ using only 2D image supervisions, volume rendering is typically utilized~\cite{wang2021neus,yariv2021volume}.
Given the sampled pixels in an image, camera rays $\{\mathbf{r}(t) = \mathbf{o} + t\mathbf{d}_{\mathrm{view}}|t>0\}$ are cast through the camera origin $\mathbf{o}$ and the sampled pixels, where $\mathbf{d}_{\mathrm{view}}$ is the unit directional vector of the ray.
To obtain the rendered colors of the sampled rays, a number of points are sampled along each ray, and queried for their SDF values and view-dependent colors from $\Theta_f$ and $\Theta_c$, where positional encoding (PE)~\cite{mildenhall2020nerf} is typically incorporated.
A relationship between the SDF value and opaque density is then formulated, such as using S-density~\cite{wang2021neus}, to translate the queried SDFs and colors to the ray color akin to NeRF~\cite{mildenhall2020nerf}.
Finally, the training loss is calculated from the difference between the rendered ray colors and the ground-truth colors of the sampled pixels.

\noindent
\textbf{Directional parameterization.}
The key to the ability of the radiance network to model view-dependent appearance lies in its incorporation of a directional vector $\mathbf{d}$ as an input.
Common directional parameterizations for $\mathbf{d}$ includes the viewing direction $\mathbf{d}_\mathrm{view}$ and reflection direction $\mathbf{d}_\mathrm{ref}$.
The viewing directional parameterization is intuitively straightforward, as it models the view-dependent appearance from the perspective of the observer.
The reflection directional parameterization is a little more complicated, which encodes the viewing direction $\mathbf{d}_\mathrm{view}$ and local normal direction $\mathbf{n}$ into a new compact directional vector $\mathbf{d}_\mathrm{ref}$, by considering the interaction between light and surface. Intuitively, it represents the direction of the incoming light, assuming that the surface is specular.
$\mathbf{d}_\mathrm{ref}$ is formulated as follows:
\begin{equation}
\label{eq:reflection}
    \mathbf{d}_{\mathrm{ref}} = 2(\mathbf{d}_{\mathrm{view}} \cdot \mathbf{n}) \mathbf{n} - \mathbf{d}_{\mathrm{view}},
\end{equation}
where $\mathbf{n}$ can be derived from the gradient of the SDF as $\mathbf{n}(\mathbf{x})= \frac{ \nabla f(\mathbf{x})} { \|\nabla f(\mathbf{x})\|}$.

\subsection{Observations and Analysis}
\label{subsec:3_2}

\noindent
\textbf{Observations.}
Although existing neural implicit surface reconstruction methods widely adopt either viewing direction or reflection direction as their directional parameterization, both types of parameterizations have their respective limitations when it comes to reconstructing certain objects.
Using viewing direction often faces difficulty in correctly reconstructing objects with specular surfaces (\eg, Fig.~\ref{fig:intro_1} (a) and Fig.~\ref{fig:method_2} (c)).
On the other hand, while using reflection direction can reconstruct specular objects with simple geometry (\eg, Fig.~\ref{fig:intro_1} (b) top and Fig.~\ref{fig:method_2} (d)), it often results in overly smooth reconstructions with missing concave surfaces when reconstructing objects with concave or complex structures (\eg, Fig.~\ref{fig:intro_1} (b) bottom and Fig.~\ref{fig:method_2} (b)).
Interestingly, using viewing direction may yield better results in these parts with complex or concave surfaces (\eg, Fig.~\ref{fig:intro_1} (a) bottom and Fig.~\ref{fig:method_2} (a)).

Based on these observations, we raise the question: Is there a unified directional parameterization that can effectively handle both specular surfaces and complex geometry?
To answer this, we start by analyzing the reasons behind their reconstruction failures:

\begin{figure}[tb]
  \centering
  \includegraphics[width=0.98\linewidth]{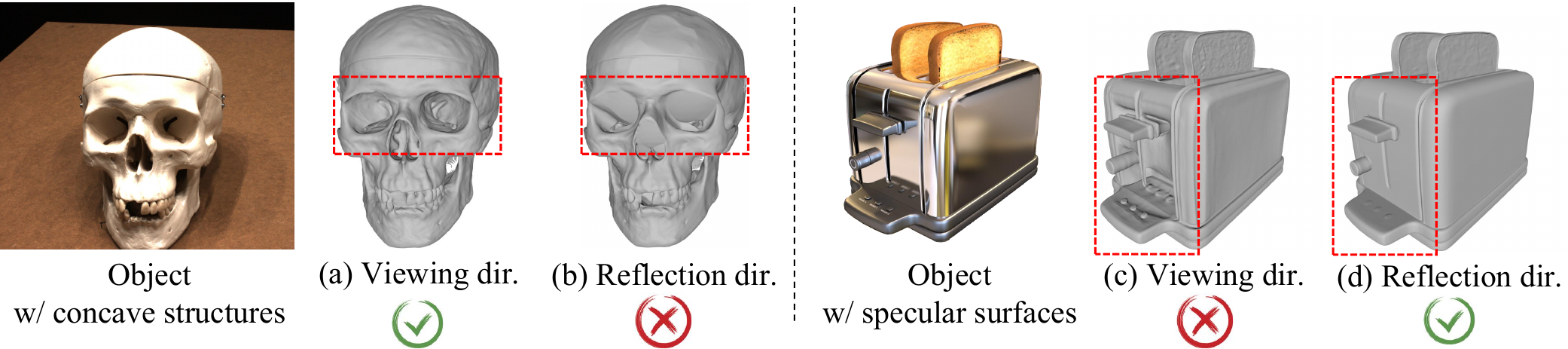}
  \caption{\textbf{Scenarios where existing directional parameterizations succeed and struggle.}
  Reconstruction results are obtained by integrating the two directional parameterizations into NeuS~\cite{wang2021neus}.
  }
  \label{fig:method_2}
\end{figure}

\noindent
\textbf{Analysis of failures when using viewing direction.}
An analysis of why using viewing direction is unsuitable for modeling the view-dependent appearance of specular objects is provided in Verbin~\etal~\cite{verbin2022ref}, which points out that the radiance network conditioned on viewing direction is poorly suited for interpolation.
We consider this characteristic leads to difficulties in correctly decoupling the geometry and appearance of specular objects when using viewing direction.
On the other hand, reflection directional parameterization can effectively leverage the prior knowledge of the interaction between light and surface, which facilitates the correct decoupling of geometry and appearance, especially for specular objects.

\noindent
\textbf{Analysis of failures when using reflection direction.}
In contrast, no work has yet provided a detailed analysis of why using reflection directional parameterization fails in the reconstruction of concave or complex structures, making us the first to provide this analysis.
We first highlight two key differences between viewing direction and reflection direction: the viewing direction is non-learnable throughout the optimization and remains constant along a sampled ray, whereas the reflection direction 1) depends on learnable geometry (\ie, normals) and 2) may vary significantly during the optimization.
We argue that these two properties of the reflection direction may pose challenges to the accurate recovery of geometry, especially for concave or complex structures.

\begin{figure}[tb]
  \centering
  \includegraphics[width=1.0\linewidth]{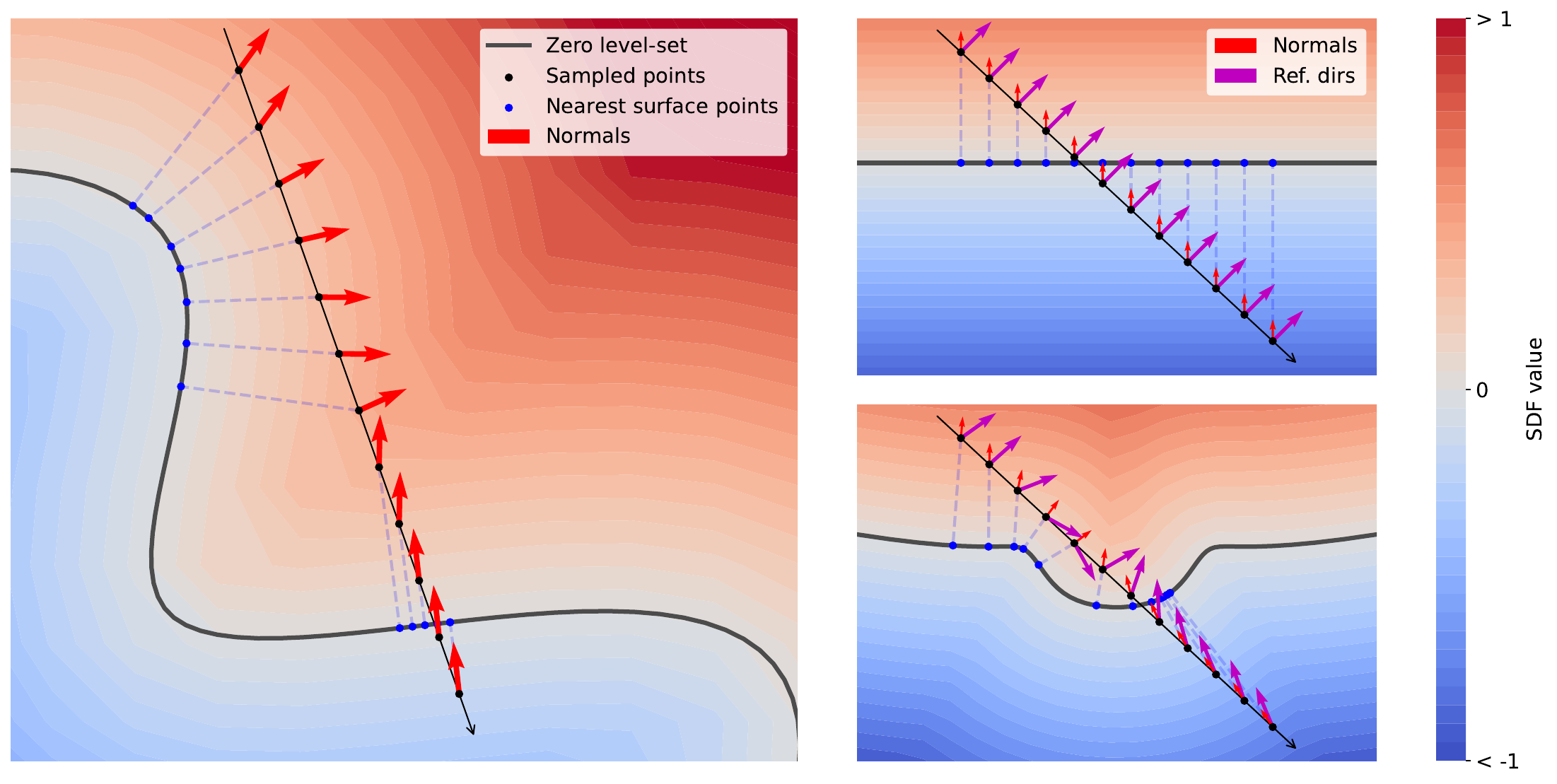}
  \caption{\textbf{Impact of incorporating normals on view-dependent radiance modeling.} \textbf{(Left)} Normals at sampled points may be influenced by unrelated surfaces other than the intersection surface. 
  \textbf{(Top right)} Smooth surfaces yield similarity in normals and reflection directions for the sampled points, while \textbf{(Bottom right)} surfaces with intricate local details, \eg, concavities, induce a scattered distribution of normals and reflection directions (Ref. dirs), particularly pronounced slightly distant from the vicinity of the zero-level set, which adversely affects geometry optimization. The details are explained in Sec.~\ref{subsec:3_2}.}
  \label{fig:method_details}
\end{figure}

A 2D toy example is first illustrated on Fig.~\ref{fig:method_details}~(left) to better demonstrate why the first property of reflection direction may hinder the recovery of geometry, which assumes the scenario of recovering an \quotes{L}-shaped concave structure.
A sampled ray intersecting the surface and the sampled points on the ray along with the normals at these points are illustrated.
In this case, we notice that the normals at sampled points that are slightly distant from the intersecting surface do not necessarily relate to the normal of the intersecting surface (the bottom side of the zero level-set), but rather to the normals of surfaces elsewhere (the left side of the zero level-set).
As a result, since the training loss of existing neural implicit surface reconstruction methods is usually ray-based, which is calculated using the local color of the intersecting surface, using this loss to optimize the normals of irrelevant surfaces elsewhere may introduce ambiguity in the optimization of geometry and result in incorrect geometric updates.

In addition, the second property of the reflection direction is more closely related to the phenomenon that using reflection directional parameterization tends to reconstruct overly smooth surfaces with missing concave or complex structures.
Assuming that the geometry reconstruction begins with a smooth surface exhibiting local flatness (Fig.~\ref{fig:method_details} top-right), which aligns with the geometric initialization practices commonly employed by existing neural implicit surface reconstruction methods~\cite{wang2021neus, Fu2022GeoNeus}, and the geometry of the target object includes concave structures (Fig.~\ref{fig:method_details} bottom-right).
We observed that for the same sampled ray, when the surface geometry is updated from a smooth surface to a concave surface, the normals and reflection directions of the same sampled points also change. This change is particularly pronounced for points that are slightly distant from the surface, and could be further pronounced due to the presence of Positional Encoding (PE).
The noticeable changes in reflection directions at these points within each surface update introduce high-frequency variances to the directional inputs of the radiance network. This makes learning a stable radiance field challenging.
To be specific, assuming that the radiance network conditioned on reflection direction has already learned to minimize the color loss to some extent before the surface update, the substantial change in directional inputs due to the update could likely lead to an increase in color loss, which hinders the optimization and is susceptible to falling into local minima.

In contrast, we notice that using the viewing direction often leads to concave or overly intricate surface details (although some may be incorrect), as shown in Fig.~\ref{fig:intro_1}~(a) and Fig.~\ref{fig:method_2}~(a) (c).
We consider that this is because the viewing direction does not rely on the surface geometry, which means that changes in geometry do not affect the optimization of the radiance network conditioned on viewing direction.
This allows the SDF can undergo more \quotes{aggressive} updates without considering the impact on radiance.

\subsection{Hybrid Directional Parameterization}
\label{subsec:3_3}
Building upon the analysis from the previous section, we discovered that viewing direction and reflection direction seem to have a spatially complementary relationship:
On one hand, for sampled points near the intersecting surface, the normals here accurately reflect the normals of the intersecting surface, thereby enabling the reflection direction to better model the interaction between light and surface, which in turn promotes the correct decomposition of geometry and appearance.
On the other hand, for other sampled points along the ray, using the reflection direction might introduce irrelevant surface geometry into the optimization, or high-frequency variations in the directional inputs of the radiance network, both potentially hindering the optimization of the geometry and radiance networks.
By contrast, using the viewing direction at these sampled points can avoid these issues.
Motivated by these thoughts, we propose a spatial-aware directional representation, which transitions from reflection direction $\mathbf{d}_{\mathrm{ref}}$  to viewing direction $\mathbf{d}_{\mathrm{view}}$ based on the distance to the surface (\ie, the absolute value of SDF).
We refer to this fused direction as \textbf{hybrid direction} $\mathbf{d}_{\mathrm{hyb}}$, formulated as:

\begin{equation}
    \label{eq:fusion}
    \mathbf{d}_\mathrm{hyb} = \mathrm{normalize}(\alpha \cdot \mathbf{d}_\mathrm{ref} + (1-\alpha) \cdot \mathbf{d}_\mathrm{view}).
\end{equation}
Here, $\alpha \in [0, 1]$ represents the blend weight. We normalize the resulting linear blend to a unit vector thereby representing a valid direction. The blend weight $\alpha$ is implemented as
\begin{equation}
    \label{eq:gamma}
    \alpha = \mathrm{exp}(-\gamma\cdot \mathrm{detach}(|f(\mathbf{\mathbf{x}})|)),
\end{equation}
where $\gamma \in \mathbb{R}^+$ is an optimizable parameter that controls the transition speed between reflection/viewing direction. 
The gradient of the SDF is detached to prevent interference with geometric updates.
At the surface of the object, \ie, when SDF equals 0, $\alpha$ simply becomes 1, which means that the hybrid direction aligns with the reflection direction. As $\mathbf{x}$ moves further away from the surface, $\alpha$ gradually approaches 0, indicating that the direction tends towards the viewing direction. 
Here, $\alpha$ can be any bell-shaped function w.r.t. the SDF value. In this case, we utilize the PDF of the Laplace distribution for simplicity.

In the next section, extensive experimental results validate the superiority of the proposed hybrid directional parameterization compared to existing parameterizations.

\section{Experiments}
\label{sec:exp}

\subsection{Experimental Settings}
\noindent
\textbf{Datasets.}
We conduct experiments on objects from different datasets, including the DTU dataset~\cite{aanaes2016large}, the Shiny Blender dataset~\cite{verbin2022ref}, and the real captured dataset~\cite{hedman2021baking}.
The DTU dataset~\cite{aanaes2016large} contains indoor multi-view captures of diverse objects, including challenging conditions for geometry reconstruction such as non-Lambertian surfaces and complex structures.
We use the same 15 objects as previous works~\cite{wang2021neus, cai2023neuda, idr, wu2022voxurf} for evaluation.
In addition, we utilize the foreground masks provided by Yariv~\etal~\cite{idr}.
The Shinny Blender dataset~\cite{verbin2022ref} is a synthetic dataset rendered in Blender~\cite{blender}, which contains different objects with specular surfaces.
We follow the experimental settings used in Ref-NeuS~\cite{ge2023ref} on this dataset, which selects four objects.
The real captured dataset~\cite{hedman2021baking} contains a set of three outdoor captures of real objects with complex geometry and specular surfaces.
We select the \quotes{toycar} scene for evaluation.

\noindent
\textbf{Implementation details.}
We integrate the proposed hybrid directional parameterization into two types of neural surface reconstruction methods: NeuS~\cite{wang2021neus} and NeuDA~\cite{cai2023neuda}.
The former utilizes a deep MLP network as the scene representation, while the latter leverages a shallow MLP decoder combined with multi-level voxel grids.
Both methods adopt a radiance network conditioned on viewing direction in their original versions.
The modification of the directional input from the viewing direction to the proposed hybrid direction can be accomplished with less than 10 lines of code.
We additionally implement the radiance network that depends on reflection direction for both methods, to facilitate a comprehensive comparison of different directional parameterizations.
Regarding the learnable parameter $\gamma$ in Eqn.~\ref{eq:gamma}, we implement it as $\gamma=\mathrm{exp}(10\cdot \gamma_b)$.
We assign an initial value of 0.3 to $\gamma_b$ except for the \quotes{toaster} scene from the Shiny Blender dataset. For the \quotes{toaster} scene, an initial value of 0.1 is assigned since it results in better reconstruction.
The training losses of the modified methods are identical to those of their original versions.
The number of training iterations is set to 300,000 to ensure convergence.

\noindent
\textbf{Evaluation metrics.}
For the DTU dataset, we follow the evaluation protocol of previous works~\cite{wang2021neus, cai2023neuda} and evaluate the Chamfer Distance (CD) between the estimated and ground-truth geometry.
For the Shiny Blender dataset, we follow the evaluation protocol proposed in Ref-NeuS~\cite{ge2023ref}. It evaluates the accuracy (Acc.) instead of the CD for unbiased evaluation, and also evaluates the mean angular error (MAE) between the ground-truth normal maps and the rendered normal maps.
For the real captured dataset~\cite{hedman2021baking}, we only provide a qualitative evaluation, as ground-truth geometry is not provided in this dataset.

\subsection{Comparison of Different Directional Parameterizations}

\begin{table}[t]
\renewcommand\arraystretch{1.1}
\caption{Quantitative evaluation in terms of Chamfer distance of integrating different directional parameterizations in NeuS~\cite{wang2021neus} and NeuDA~\cite{cai2023neuda} on the DTU dataset~\cite{aanaes2016large}.
The experimental objects are categorized into diffuse objects and specular objects.
The numbers in the headers denote the scene IDs.
\textbf{Bold} results have the best score.}
\label{tab:01}
\resizebox{\textwidth}{!}{%
\begin{tabular}{ll|cccccccccccc|ccccc|c}
\hline
\multicolumn{2}{l|}{\multirow{2}{*}{Methods}} & \multicolumn{12}{c|}{Scan of Diffuse Objects}                                                                                                                                                 & \multicolumn{5}{c|}{Scan of Specular Objects}                                 & \multirow{2}{*}{\begin{tabular}[c]{@{}c@{}}Total\\ Avg.\end{tabular}} \\ \cline{3-19}
\multicolumn{2}{l|}{}                         & 24            & 37            & 40            & 55            & 65            & 83            & 105           & 106           & 114           & 118           & 122           & Avg.          & 63            & 69            & 97            & 110           & Avg.          &                                                                       \\ \hline
\multicolumn{2}{l|}{NeuS~\cite{wang2021neus}}                     & 0.83          & 0.98          & 0.56          & \textbf{0.37} & \textbf{0.59} & \textbf{1.45} & 0.78          & 0.52          & 0.36          & \textbf{0.45} & \textbf{0.45} & \textbf{0.67} & 1.13          & \textbf{0.60} & 0.95          & 1.43          & 1.03          & 0.77                                                                  \\
\multicolumn{2}{l|}{NeuS, w/ reflection dir.}  & 1.39          & 4.39          & 0.73          & 0.39          & 1.34          & 1.49          & 0.92          & 0.59          & 0.36          & 0.46          & 0.54          & 1.14          & 1.11          & 0.63          & 1.46          & 0.92          & 1.03          & 1.11                                                                  \\
\multicolumn{2}{l|}{NeuS, w/ hybrid dir. (Ours)}      & \textbf{0.81} & \textbf{0.97} & \textbf{0.54} & 0.38          & 0.61          & 1.47          & \textbf{0.77} & \textbf{0.51} & \textbf{0.35} & \textbf{0.45} & 0.47          & \textbf{0.67} & \textbf{1.10} & \textbf{0.60} & \textbf{0.85} & \textbf{0.90} & \textbf{0.86} & \textbf{0.72}                                                         \\ \hline
\multicolumn{2}{l|}{NeuDA~\cite{cai2023neuda}}                    & \textbf{0.51} & \textbf{0.76} & \textbf{0.39} & 0.37          & \textbf{0.56} & \textbf{1.37} & 0.79          & 0.50          & 0.34          & 0.42          & \textbf{0.46} & \textbf{0.59} & 1.08          & 0.57          & 1.13          & 0.80          & 0.90          & 0.67                                                                  \\
\multicolumn{2}{l|}{NeuDA, w/ reflection dir.} & 0.58          & 1.03          & 0.63          & 0.37          & 0.79          & 1.45          & 0.89          & 0.51          & 0.37          & 0.43          & 0.50          & 0.69          & 1.03          & 0.55          & 0.90          & 0.76          & 0.81          & 0.72                                                                  \\
\multicolumn{2}{l|}{NeuDA, w/ hybrid dir. (Ours)}     & 0.53          & 0.77          & 0.40          & \textbf{0.36} & 0.57          & 1.38          & \textbf{0.76} & \textbf{0.47} & \textbf{0.33} & \textbf{0.41} & 0.47          & \textbf{0.59} & \textbf{1.02} & \textbf{0.54} & \textbf{0.76} & \textbf{0.75} & \textbf{0.77} & \textbf{0.63}                                                         \\ \hline
\end{tabular}
}
\end{table}

\begin{figure}[t]
  \centering
  \includegraphics[width=0.98\linewidth]{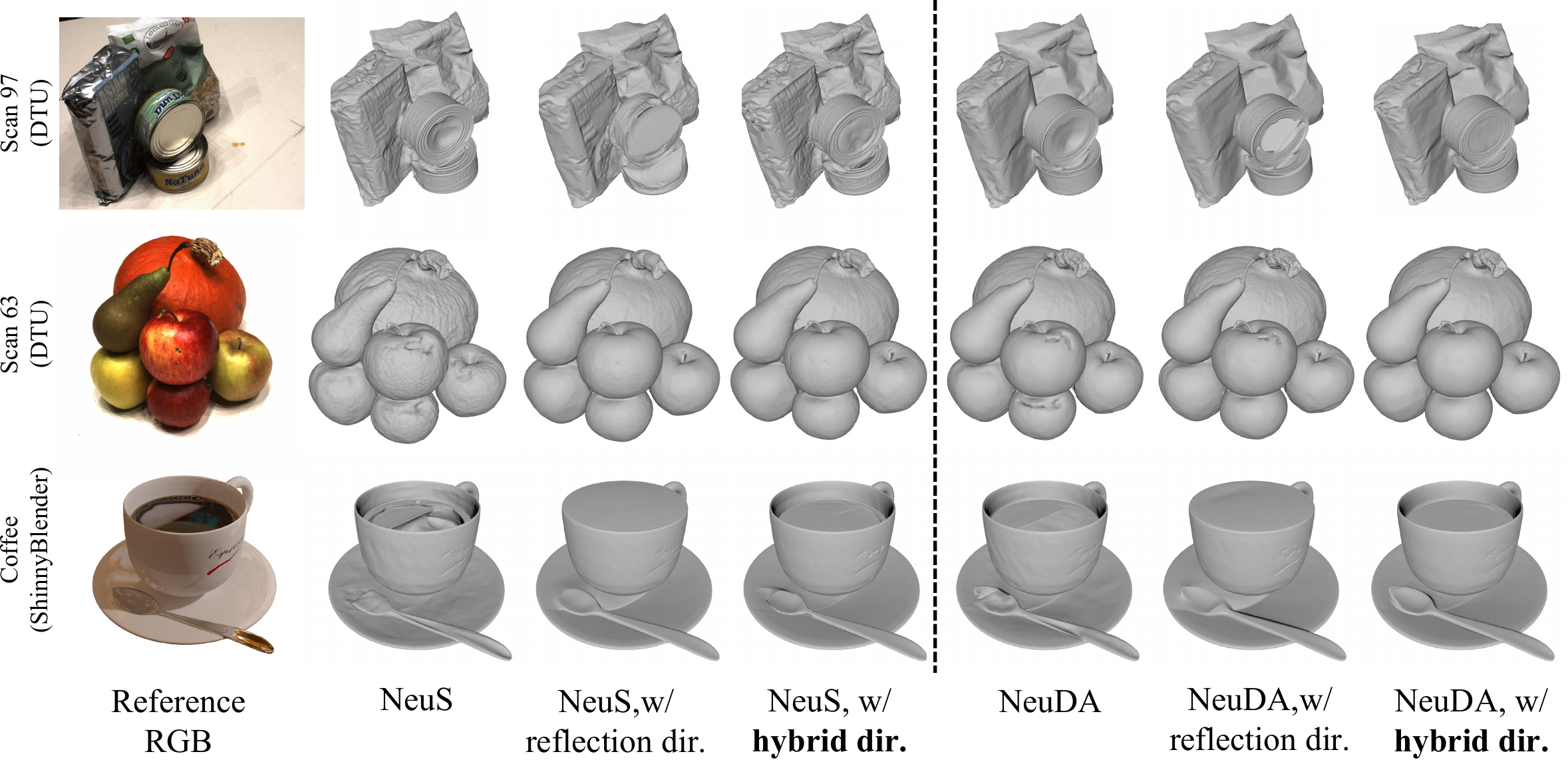}
  \caption{Qualitative comparison of integrating different directional parameterizations in NeuS~\cite{wang2021neus} and NeuDA~\cite{cai2023neuda} on the DTU dataset~\cite{aanaes2016large} and Shiny Blender dataset~\cite{verbin2022ref}.}
  \label{fig:exp_1}
\end{figure}

\noindent
We first evaluate the performance of integrating different directional parameterizations into the two baseline methods NeuS~\cite{wang2021neus} and NeuDA~\cite{cai2023neuda} on the DTU dataset~\cite{aanaes2016large}.
The experimental objects from the DTU dataset are categorized into diffuse objects and specular objects for better demonstration.
Tab.~\ref{tab:01} summarizes the quantitative results.
For both baseline methods, we notice that substituting the original viewing directional parameterization with the reflection direction only yields better results for a minority of specular objects.
For diffuse objects and the remaining specular objects, employing reflection directional parameterization demonstrates significant degradation in performance.
Considering the relatively complex geometry of DTU objects, we attribute this to the optimization issue analyzed in Sec.~\ref{subsec:3_2}.
Conversely, the incorporation of the proposed hybrid directional parameterization enables both modified methods to yield consistently superior performance for specular objects while achieving results on par with the original version for diffuse objects.
The first two rows in Fig.~\ref{fig:exp_1} showcase the visual comparison of adopting different directional parameterizations for the reconstruction of two specular objects from the DTU dataset.
Both baseline methods fail to reconstruct the specular surfaces of objects.
On the other hand, although integrating the reflection direction yields better reconstruction for specular objects with simple geometry (\eg, Scan 63), they fail to reconstruct the specular surfaces at the concave structure in Scan 97.
In contrast, integrating the proposed hybrid direction achieves the best reconstruction results in both scenarios.
More qualitative comparisons on the DTU dataset are available in our supplementary.

\begin{table}[t]
\renewcommand\arraystretch{1.1}
\caption{Quantitative evaluation of integrating different directional parameterizations in NeuS~\cite{wang2021neus} and NeuDA~\cite{cai2023neuda} on the Shiny Blender dataset~\cite{verbin2022ref}.
\textbf{Bold} results have the best score.}
\label{tab:02}
\centering
\resizebox{1\textwidth}{!}{%
\begin{tabular}{ll|cc|cc|cc|cc|cc}
\hline
\multicolumn{2}{l|}{\multirow{2}{*}{Methods}} & \multicolumn{2}{c|}{helmet} & \multicolumn{2}{c|}{toaster} & \multicolumn{2}{c|}{coffee} & \multicolumn{2}{c|}{car} & \multicolumn{2}{c}{Avg.} \\ \cline{3-12} 
\multicolumn{2}{l|}{}            & Acc. $\downarrow$         & MAE $\downarrow$          & Acc. $\downarrow$          & MAE $\downarrow$          & Acc. $\downarrow$         & MAE $\downarrow$          & Acc. $\downarrow$        & MAE $\downarrow$        & Acc. $\downarrow$        & MAE $\downarrow$        \\ \hline
\multicolumn{2}{l|}{NeuS~\cite{wang2021neus}}                     & 4.88         & 3.20         & 3.31          & 2.85         & 1.97         & 1.06         & 0.86        & 0.95       & 2.76        & 2.02       \\
\multicolumn{2}{l|}{NeuS, w/ reflection dir.}  & 0.31         & 0.36         & \textbf{0.39}          & 0.94         & 3.30         & 1.37         & 0.48        & 0.73       & 1.12        & 0.85       \\
\multicolumn{2}{l|}{NeuS, w/ hybrid dir. (Ours)}      & \textbf{0.29}         & \textbf{0.34}         & 0.45          & \textbf{0.92}         & \textbf{0.64}         & \textbf{0.55}         & \textbf{0.43}        & \textbf{0.72}       & \textbf{0.45}        & \textbf{0.63}       \\ \hline
\multicolumn{2}{l|}{NeuDA~\cite{cai2023neuda}}                    & 5.73             & 3.77         & 5.98              & 3.39         & 1.16             & 0.81         & 0.87            & 1.03       & 3.44            & 2.25       \\
\multicolumn{2}{l|}{NeuDA, w/ reflection dir.} & 0.34             & 0.37         & 1.03              & 1.71         & 3.66             & 1.53         & 0.62            & 0.87       & 1.41            & 1.12       \\
\multicolumn{2}{l|}{NeuDA, w/ hybrid dir. (Ours)}     & \textbf{0.32}             & \textbf{0.36}         & \textbf{0.95}              & \textbf{1.54}         & \textbf{0.55}             & \textbf{0.57}         & \textbf{0.60}            & \textbf{0.84}       & \textbf{0.61}            & \textbf{0.83}       \\ \hline
\end{tabular}
}
\end{table}

We further validate the advantage of the proposed hybrid directional parameterization on the Shiny Blender dataset~\cite{verbin2022ref}, as demonstrated in Tab.~\ref{tab:02}.
Both baseline methods exhibit large reconstruction errors on the surfaces of highly specular objects in this dataset.
By contrast, due to the relatively simple geometry of objects, replacing the viewing directional parameterization with the reflection direction results in a significant improvement in reconstruction quality on most objects in this dataset, except for the \quotes{coffee} object.
For the reconstruction of the \quotes{coffee} object, adopting the reflection directional parameterization fails to recover the concave structure formed by the inner wall of the coffee cup and the surface of the water, as shown in the third row in Fig.~\ref{fig:exp_1}.
On the other hand, the integration of our proposed hybrid directional parameterization enables the successful reconstruction of both the specular surfaces and concave structures, obtaining consistently superior reconstructions of all objects in this dataset, compared to the other two directional parameterizations.

\begin{figure}[t]
  \centering
  \includegraphics[width=0.98\linewidth]{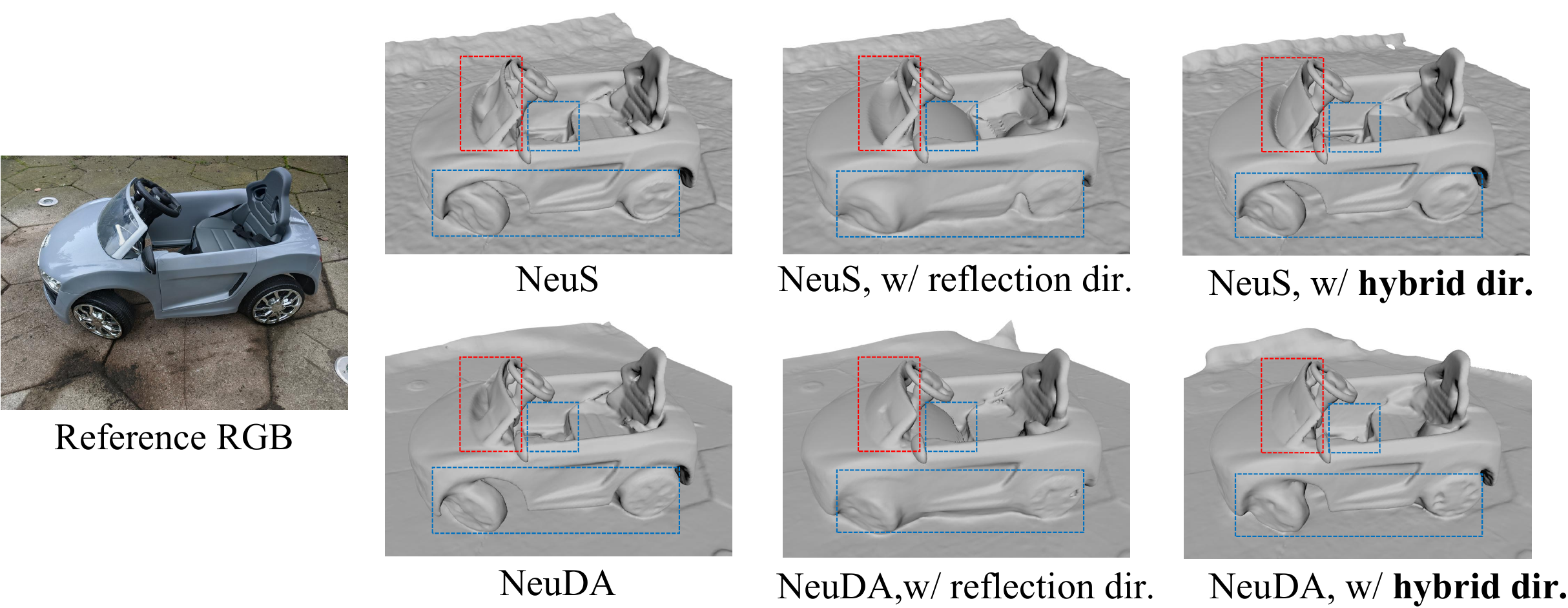}
  \caption{Qualitative comparison of integrating different directional parameterizations in NeuS~\cite{wang2021neus} and NeuDA~\cite{cai2023neuda} on the \quotes{toycar} scene from the real captured dataset~\cite{hedman2021baking}.}
  \label{fig:exp_2}
\end{figure}

Lastly, we qualitatively present the reconstruction results on the \quotes{toycar} scene from the real captured scene in Fig.~\ref{fig:exp_2}, which contains challenging reflective glass and complex geometry.
Both modified methods employing the proposed hybrid directional parameterization succeed in recovering the complex geometry and the smooth surface of the reflective glass.
In contrast, using either the reflection direction or the original viewing direction results in reconstructions that are either overly smooth, missing concave surface structures, or provide incorrect surface reconstructions at the reflective glass areas.

As the summary of this section, compared to the viewing directional parameterization which struggles with reconstructing specular surfaces, and the reflection directional parameterization which faces difficulties in reconstructing complex structures, our proposed hybrid directional parameterization effectively handles both scenarios in a unified form.
This benefit is attributed to our hybrid directional parameterization effectively combining the advantages of both viewing and reflection directional parameterizations while avoiding their drawbacks.

\subsection{Comparison to State-of-the-Art}

Although our main purpose is not to develop a best-performing surface reconstruction system, we demonstrate that we can still achieve state-of-the-art performance by integrating the proposed hybrid directional parameterization into existing methods.
For brevity, we refer to NeuS~\cite{wang2021neus} and NeuDA~\cite{cai2023neuda} incorporating the proposed hybrid directional parameterization as Ours (NeuS) and Ours (NeuDA) in the following text.
Tab.~\ref{tab:03} reports the Chamfer distance (CD) metrics of different methods on the DTU dataset~\cite{aanaes2016large}.
For a fair comparison, all methods in Tab.~\ref{tab:03} utilize foreground masks provided by Yariv~\etal~\cite{idr} during training.
Both Ours (NeuS) and Ours (NeuDA) improve their baselines by a certain margin in the average CD scores, and Ours (NeuDA) achieves the best performance compared to previous SOTA methods in the average CD metric.

\begin{table}[t]
\renewcommand\arraystretch{1.1}
\caption{Quantitative evaluation in terms of Chamfer distance on the DTU dataset~\cite{aanaes2016large}.
The numbers in the header denote the scene IDs.
\textbf{Bold} results have the best score.}
\label{tab:03}
\centering
\resizebox{1\textwidth}{!}{%
\begin{tabular}{l|ccccccccccccccc|c}
\hline
Methods      & 24                       & 37                       & 40                       & 55                       & 63                       & 65                       & 69                       & 83                       & 97                       & 105                      & 106                      & 110                      & 114                      & 118                      & 122                       & Avg.                     \\ \hline
IDR~\cite{idr}          & 1.63                     & 1.87                     & 0.63                     & 0.48                     & 1.04                     & 0.79                     & 0.77                     & \textbf{1.33}            & 1.16                     & 0.76                     & 0.67                     & 0.90                     & 0.42                     & 0.51                     & 0.53                      & 0.90                     \\
NeuS~\cite{wang2021neus}         & 0.83                     & 0.98                     & 0.56                     & 0.37                     & 1.13                     & 0.59                     & 0.60                     & 1.45                     & 0.95                     & 0.78                     & 0.52                     & 1.43                     & 0.36                     & 0.45                     & \textbf{0.45}             & 0.77                     \\
Voxurf~\cite{wu2022voxurf}       & 0.65                     & \textbf{0.74}            & \textbf{0.39}            & \textbf{0.35}            & \textbf{0.96}            & 0.64                     & 0.85                     & 1.58                     & 1.01                     & \textbf{0.68}            & 0.60                     & 1.11                     & 0.37                     & 0.45                     & 0.47                      & 0.72                     \\
NeuDA~\cite{cai2023neuda}        & \textbf{0.51}            & 0.76                     & \textbf{0.39}            & 0.37                     & 1.08                     & \textbf{0.56}            & 0.57                     & 1.37                     & 1.13                     & 0.79                     & 0.50                     & 0.80                     & 0.34                     & 0.42                     & 0.46                      & 0.67                     \\ \hline
Ours (NeuS)  & \multicolumn{1}{l}{0.81} & \multicolumn{1}{l}{0.97} & \multicolumn{1}{l}{0.54} & \multicolumn{1}{l}{0.38} & \multicolumn{1}{l}{1.10} & \multicolumn{1}{l}{0.61} & \multicolumn{1}{l}{0.60} & \multicolumn{1}{l}{1.47} & \multicolumn{1}{l}{0.85} & \multicolumn{1}{l}{0.77} & \multicolumn{1}{l}{0.51} & \multicolumn{1}{l}{0.90} & \multicolumn{1}{l}{0.35} & \multicolumn{1}{l}{0.45} & \multicolumn{1}{l|}{0.47} & \multicolumn{1}{l}{0.72} \\
Ours (NeuDA) & 0.53                     & 0.77                     & 0.40                     & 0.36                     & 1.02                     & 0.57                     & \textbf{0.54}            & 1.38                     & \textbf{0.76}            & 0.76                     & \textbf{0.47}            & \textbf{0.76}            & \textbf{0.33}            & \textbf{0.41}            & 0.47                      & \textbf{0.63}            \\ \hline
\end{tabular}
}
\end{table}

Tab.~\ref{tab:04} reports the quantitative evaluation of different methods on the Shiny Blender dataset~\cite{verbin2022ref}.
The results of ~\cite{verbin2022ref, unisurf, neuralwarp, Fu2022GeoNeus, ge2023ref, yariv2021volume} are taken from the results reported in Ge~\etal~\cite{ge2023ref}.
Both Ours (NeuS) and Ours (NeuDA) achieve better average MAE metrics than the previous SOTA method Ref-NeuS~\cite{ge2023ref}, and Ours (NeuS) achieve the best overall performance in terms of both Acc. and MAE metrics on this dataset.
We refer to our supplementary for the qualitative comparisons between Ours (NeuS) and Ref-NeuS~\cite{ge2023ref}.

\begin{table}[t]
\renewcommand\arraystretch{1.1}
\caption{Quantitative evaluation on the Shiny Blender dataset~\cite{verbin2022ref}.
\quotes{-} denotes that the results are not available.
\textbf{Bold} results have the best score.}
\label{tab:04}
\centering
\resizebox{0.8\textwidth}{!}{%
\begin{tabular}{l|cc|cc|cc|cc|cc}
\hline
\multirow{2}{*}{Methods} & \multicolumn{2}{c|}{helmet} & \multicolumn{2}{c|}{toaster}  & \multicolumn{2}{c|}{coffee}   & \multicolumn{2}{c|}{car}      & \multicolumn{2}{c}{Avg.}      \\ \cline{2-11} 
                         & Acc. $\downarrow$     & MAE $\downarrow$             & Acc. $\downarrow$         & MAE $\downarrow$          & Acc. $\downarrow$         & MAE $\downarrow$          & Acc. $\downarrow$         & MAE $\downarrow$          & Acc. $\downarrow$         & MAE $\downarrow$          \\ \hline
Ref-NeRF~\cite{verbin2022ref}                 & -        & 29.48            & -             & 42.87         & -             & 12.24         & -             & 14.93         & -             & 24.88         \\
NeuS~\cite{wang2021neus}                     & 4.88     & 3.20             & 3.31          & 2.85          & 1.97          & 1.06          & 0.86          & 0.95          & 2.76          & 2.02          \\
UNISURF~\cite{unisurf}                  & 1.69     & 1.78             & 3.75          & 5.51          & 2.88          & 3.15          & 2.34          & 1.98          & 2.67          & 3.11          \\
VolSDF~\cite{yariv2021volume}                   & 1.55     & 1.37             & 2.02          & 2.53          & 2.23          & 2.28          & 0.58          & 1.04          & 1.60          & 1.81          \\
NeuralWarp~\cite{neuralwarp}               & 2.25     & 1.94             & 5.90          & 3.51          & 1.54          & 2.04          & 0.65          & 1.07          & 2.59          & 2.14          \\
Geo-NeuS~\cite{Fu2022GeoNeus}                 & 0.74     & 2.36             & 6.35          & 3.76          & 3.85          & 6.36          & 2.88          & 5.67          & 3.46          & 4.54          \\
NeuDA~\cite{cai2023neuda}                    & 5.73         & 3.77             & 5.98              & 3.39          & 1.16              & 0.81          & 0.87              & 1.03          & 3.44              & 2.25          \\
Ref-NeuS~\cite{ge2023ref}                 & \textbf{0.29}     & 0.38             & \textbf{0.42} & 1.47          & 0.77          & 0.99          & \textbf{0.37} & 0.80          & 0.46          & 0.91          \\ \hline
Ours (NeuS)              & \textbf{0.29}     & \textbf{0.34}    & 0.45          & \textbf{0.92} & 0.64 & \textbf{0.55} & 0.43          & \textbf{0.72} & \textbf{0.45} & \textbf{0.63} \\
Ours (NeuDA)             & 0.32         & 0.36             & 0.95              & 1.54          & \textbf{0.55}              & 0.57          & 0.60              & 0.84          & 0.61              & 0.83          \\ \hline
\end{tabular}
}
\end{table}

\subsection{Ablation Studies}
In this section, we conduct ablation studies on several key design choices of the proposed hybrid directional parameterization by using Ours (NeuDA).

\begin{figure}[tb]
  \centering
  \includegraphics[width=1\linewidth]{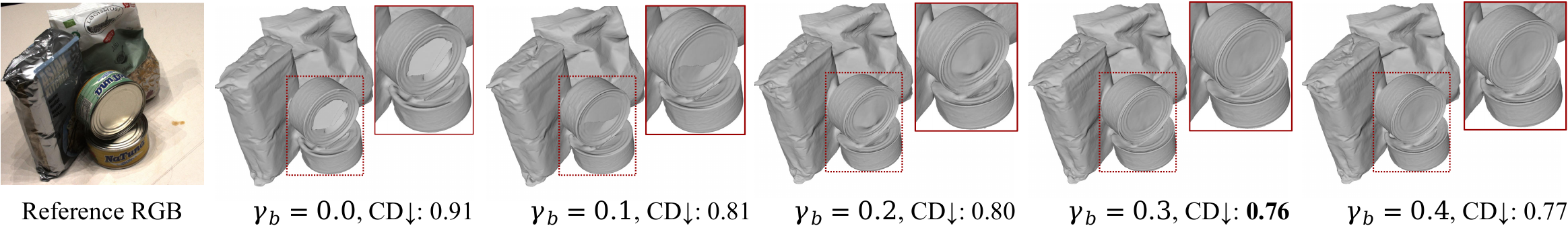}
  \caption{Reconstruction results using different initial values of $\gamma_b$.
  The main differences are highlighted in red boxes.
  }
  \label{fig:ablation1}
\end{figure}

\noindent
\textbf{The initial value of $\gamma_b$.}
We first validate the impact of setting different initial values for $\gamma_b$ on the reconstruction results.
Fig.~\ref{fig:ablation1} compares the obtained reconstruction results using five different initial values for $\gamma_b$.
We find that when the initial value of $\gamma_b$ is set low, our method tends to recover incorrect geometries similar to those caused by using the reflection directional parameterization.
This phenomenon remains consistent with our analysis in Sec~\ref{subsec:3_2}, as a smaller initial value for $\gamma_b$ increases the likelihood of involving irrelevant surfaces and high-frequency variations in the directional inputs of the radiance network during the optimization.
We empirically find the best initial value of $\gamma_b$ to be 0.3 for most experimental objects.
More discussions regarding the parameter $\gamma_b$ and blend weight $\alpha$ can be found in our supplementary material.

\begin{figure}[tbp]
\centering
\begin{minipage}[t]{0.49\linewidth}
  \adjustbox{valign=t}{\includegraphics[width=\linewidth]{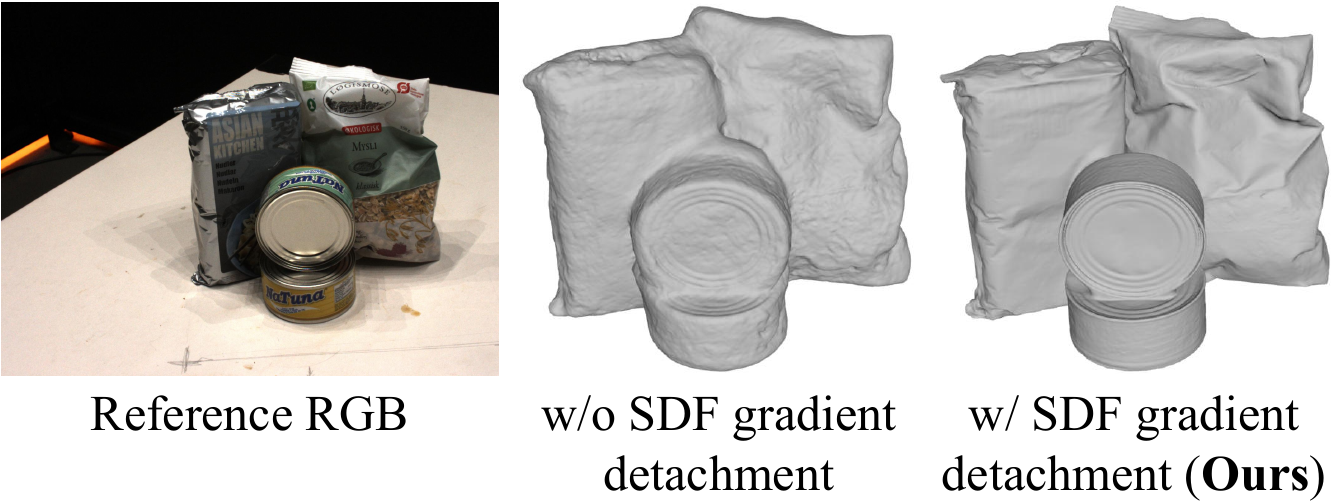}}
  \caption{Reconstruction results with and without detaching the gradient of the estimated SDF in Eqn.~\ref{eq:gamma}.}
  \label{fig:ablation2}
\end{minipage}%
\hfill
\begin{minipage}[t]{0.49\linewidth}
  \adjustbox{valign=t}{\includegraphics[width=\linewidth]{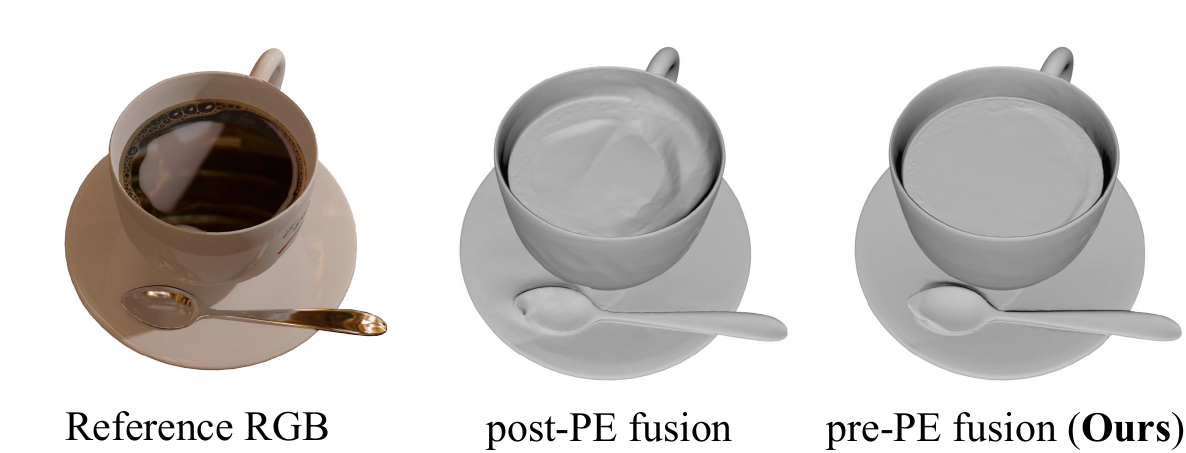}}
  \caption{Reconstruction results fusing directions before and after positional encoding (PE).}
  \label{fig:ablation3}
\end{minipage}
\end{figure}

\noindent
\textbf{Detach SDF gradient in Eqn.~\ref{eq:gamma} or not?}
Fig.~\ref{fig:ablation2} compares the reconstruction results with and without detaching the gradient of the estimated SDF in Eqn.~\ref{eq:gamma}.
We observe that our method struggles to reconstruct the fine-grained geometry when the gradient of the estimated SDF is not detached in Eqn.~\ref{eq:gamma}.
We consider that introducing the gradient of the estimated SDF places an additional burden on the learning of both the geometry and radiance networks, making it easier to fall into local minima.
Thereby, we choose to detach the gradient of SDF in Eqn.~\ref{eq:gamma}.

\noindent
\textbf{Fuse the viewing direction and reflection direction before or after positional encoding?}
We introduce the fusion of viewing direction and reflection direction in Eqn.~\ref{eq:fusion} before the operation of position encoding (PE).
We compare the reconstruction results with different fusion orders, specifically the proposed pre-PE fusion: $\mathrm{PE}(\mathbf{d}_\mathrm{hyb})$ (ours) and alternative approach using post-PE fusion: $\alpha \cdot\mathrm{PE}(\mathbf{d}_\mathrm{ref}) + (1-\alpha) \cdot \mathrm{PE}(\mathbf{d}_\mathrm{view})$.
The comparison is shown in Fig.~\ref{fig:ablation3}.
We observe that fusing directions after the PE operation may result in degraded reconstruction results.
We consider that fusing the directional information in the high-dimension space may increase the complexity of the learning space of the radiance network, which brings challenges in finding optimal reconstruction results.
Thereby, we choose to fuse the viewing direction and reflection direction before the operation of positional encoding.

\section{Conclusion and Limitation}
\label{sec:conclusion}
In this paper, we conduct an in-depth investigation into the topic of directional parameterization in neural implicit surface reconstruction.
We reveal that the commonly used directional parameterizations in modeling the view-dependent radiance field, such as viewing and reflection directions, both have their limitations and none of them can effectively handle both specular surfaces and complex structures.
A novel hybrid directional parameterization is introduced to achieve this goal, which softly fuses the viewing and reflection directions based on the distance of the sampled points from the surface, uniting the advantages of each representation.
Extensive experiments demonstrate that our hybrid directional parameterization outperforms the existing directional parameterizations in reconstructing objects with a wide variety of materials, geometry, and appearance.

One limitation of our method is that the optimal choice of initial value for $\gamma_b$ may vary depending on the target object, which would require additional efforts for parameter tuning.
An automated method for selecting the optimal initial value of $\gamma_b$ for target objects could be a potential future work for us.

\newpage
\bibliographystyle{splncs04}
\bibliography{main}
\end{document}